# Facilitating automated conversion of scientific knowledge into scientific simulation models with the Machine Assisted Generation, Calibration, and Comparison (MAGCC) Framework

Chase Cockrell[1], Scott Christley[2], Gary An[1]

1. Department of Surgery, University of Vermont Larner College of Medicine
2. Department of Clinical Sciences/ Division of Bioinformatics, University of Texas, Southwestern Medical Center

**Abstract**

The Machine Assisted Generation, Comparison, and Calibration (MAGCC) framework provides machine assistance and automation of recurrent crucial steps and processes in the development, implementation, testing, and use of scientific simulation models. MAGCC bridges systems for knowledge extraction via natural language processing or extracted from existing mathematical models and provides a comprehensive workflow encompassing the composition of scientific models and artificial intelligence (AI)-assisted code generation. MAGCC accomplishes this through: 1) the development of a comprehensively expressive formal knowledge representation knowledgebase, the Structured Scientific Knowledge Representation (SSKR) that encompasses all the types of information needed to make any simulation model, 2) the use of an artificially-intelligent logic-reasoning system, the Computational Modeling Assistant (CMA), that takes information from the SSKR and generates, in a traceable fashion, model specifications across a range of simulation modeling methods, and 3) the use of the CMA to generate compliable/executable code for a simulation model from those model specifications. The current MAGCC framework can be customized any scientific domain's specific knowledgebase and existing mathematical/computational models, and future work will involve expanding the types of computational model representation that can be generated and integrating newly-developed code generating AI systems.

## 1.0 Introduction

Given the increasingly ubiquitous use of simulation models in nearly every aspect of science, there has been considerable interest in facilitating the development, enhancement and re-use of such models through automation. Central to this task is the ability to formally represent domain knowledge in a format that can be operated over/computed. While in some research domains, such as particle physics, domain knowledge can be encoded and stored primarily in mathematical equations, other fields, such as epidemiology and biomedicine, there exists no compact and formal representation of domain knowledge. The most comprehensive representations of biomedical knowledge (with knowledge as the potential progeny of data and mechanistic/causal theory) are computational models, which themselves are dynamic representations of knowledge harvested from rigorous and reproducible experimentation.

Furthermore, in these fields without formal knowledge representation formats, what simulation models do exist are often created by individual researchers/groups on an ad hoc basis, with little initial regard to model sharing, external enhancement or interoperability with other models (see [1] for a discussion on the challenges that face the biomedical field in this critical area). While these ad hoc models may prove useful for their developers, the limits of the resulting intellectual "islands" become very evident if, for some reason, expansion or repurposing of these models becomes a critical capability (for example, see the plethora of epidemiological models utilized, with extremely variable and generally inadequate performance, in the initial response to the COVID-19 pandemic).

To once again contrast with physics, the knowledge in, for example, nuclear structure models [2, 3], is conserved across models – no one doubts that the solution of the problem is the solution of the many-body Schrodinger equation; however, the manner in which the equation is solved (choice of solver, selection of cutoff/truncation point) can affect the results. In biomedical models, two models attempting to address the same problem will incorporate distinct mechanistic knowledge (of varying provenance), which makes comparison between models difficult.

Having the ability to rapidly and efficiently evaluate the knowledge and assumptions present in a scientific simulation model, to identify what sort of data is needed to link that model to the real world, to determine what sort of questions the model is able to answer, and to potentially expand/extend that model to answer new questions of interest, would be a critical capability in a whole host of areas where scientific investigation intersects with public policy. Such areas include: epidemic/pandemic response, natural resource/ecosystem management, geopolitical strategy, drug discovery/design, natural disaster response, etc.

We further note that these challenges regarding refinement, reuse and repurposing are often characterized by a significantly higher degree of epistemic uncertainty regarding the causal mechanisms which drive complex system behavior, as opposed to industrial applications, where processes are rigorously defined and there exist robust automated/semiautomated simulation development tools. As such, we see two primary differences between the industrial and scientific simulation that limit the direct expansion of industrial simulation development tools. Firstly, the operational efficiencies desired in industry lead to rapid adoption and implementation of process standards, providing a greater uniformity of what tasks simulations are expected to address; as a result the generalizable expressiveness required of automated systems is much more limited and thus more readily tractable. This is the opposite case in scientific simulations, which are exploratory almost by definition; as such, attempts to create generalizable solutions are essentially bounded by human imagination. Secondly, the distributed nature of scientific investigation, and their computational approaches, leads to customized solutions tailored to very specific scientific questions, often implemented by individuals without formal training in developing robust and sustainable software/models (and where "sharing" models can be viewed as giving up intellectual/academic capital). These two features: generalizability and customization, are intrinsically at odds, but are equally important if scientific simulation is to strive for industrial-level performance.

This paper introduces the Machine Assisted Generation, Composition and Calibration (MAGCC) framework for facilitating the semi-automated conversion of domain knowledge into executable simulations. Our project aims to fill a gap that we believe is underappreciated in the process of automating the process of going from unstructured or opaquely structured knowledge to executable simulations that can be used for specific simulation experiments. Research and advances in natural language processing (NLP) have long striven to be able to extract domain specific knowledge from text, which is the most common repository of scientific knowledge [4]. We acknowledge the extensive expertise and accomplishments in this area and rather than trying to duplicate efforts in NLP we instead focus on the next question inherent in the transition of text to simulation model: what knowledge is required to be extracted and potentially added to in for a/any simulation model to be constructed? Therefore, the first component of the MAGCC framework is a knowledgebase, the Structured Scientific Knowledge Representation (SSKR) object, that serves as a "target" for the output of automated NLP knowledge extraction applications. The SSKR in intended to encompass all the domain specific information needed to create any type of simulation model. Because the information needed to create "any" type of model may not be available for a specific use case, the next component of MAGCC is the Computational Modeling Assistant (CMA), which is a logic-based reasoning/artificial intelligence agent that determines what types of simulation models can be composed from the provided information populated in the SSKR. For this task, the CMA will output model specifications for all the modeling methods that can be created given a SSKR. Next, the CMA generates executable code for each of those model specifications, producing a first-pass simulation model that can be run and interrogated. Finally, MAGCC includes a suite of machine learning (ML) tools to allow for the calibration of the simulation model to existing data and comparison between variants of simulation models regarding use-cases in a process termed Model Exploration (ME); this component of MAGCC is termed the Machine Learning Model Exploration (MLME) environment. The design of MAGCC is intended to be broadly applicable through the comprehensive expressiveness and generalizability of its components (the SSKR, the CMA and MLME), which can also be tailored to allow customization for specific tasks through their extensibility. An overview of the MAGCC framework can be seen in Figure 1. Specific details of the various components of MAGCC are provided in the sections below.

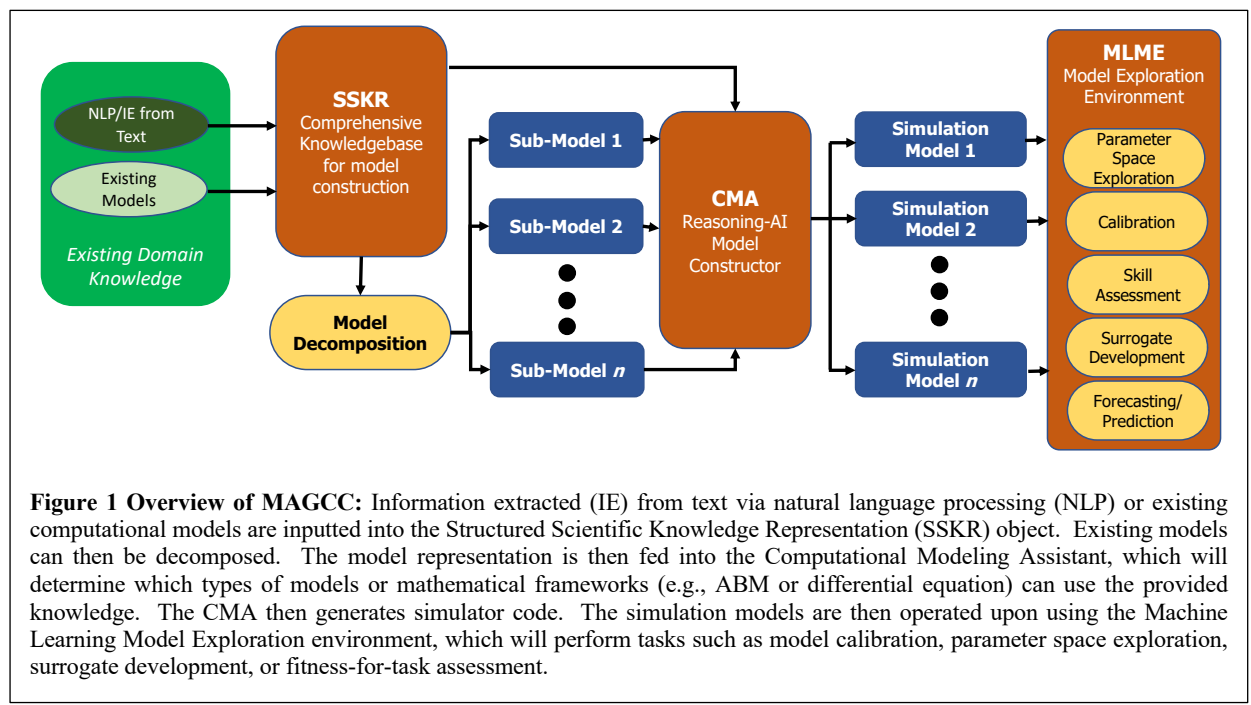

**Figure 1 Overview of MAGCC:** Information extracted (IE) from text via natural language processing (NLP) or existing computational models are inputted into the Structured Scientific Knowledge Representation (SSKR) object. Existing models can then be decomposed. The model representation is then fed into the Computational Modeling Assistant, which will determine which types of models or mathematical frameworks (e.g., ABM or differential equation) can use the provided knowledge. The CMA then generates simulator code. The simulation models are then operated upon using the Machine Learning Model Exploration environment, which will perform tasks such as model calibration, parameter space exploration, surrogate development, or fitness-for-task assessment.

## 2.0 Methods and Results

*2.1: Generalizable Model Representation for Model Composition, Decomposition and Extension: The Structured Scientific Knowledge Representation (SSKR)*

We have designed the SSKR such that it is comprehensive, meaning that the SSKR representation contains information sufficient to both define a model and implement it into an executable simulation model (through the generation of computer code) that can be operated on reliably for prognostic or diagnostic inference. We note that every element of the SSKR is not necessary or required for all models or all model use-cases, however the full set of information placed into the SSKR allows the model specification to be maximally generalizable over the potential mathematical/computational frameworks into which the model would ultimately be implemented. The SSKR has been designed such that it can sufficiently represent any system that can be modelled computationally. The SSKR is designed to interface with any system that purports to extract scientific knowledge from some pre-existing object: in some cases, the input to the SSKR will be new knowledge extracted from the literature or domain expertise via an external NLP system, in other cases it will be knowledge extracted from existing simulation models. The SSKR comprises five components.

1. Model Rule Matrix (MRM): The set of model variables and the relevant interactions between them. The Model Rule Matrix (MRM) is a formal mathematical object that contains all the variables present in a potential model in a matrix[5]. In the MRM formalism, individual rows in the matrix represent 'rules,' i.e., how features of interest in the model are updated. The columns of the matrix represent all variables in the model that contribute to the functional purpose of the model (i.e., a variable that has the purpose of indicating what should be written to file at the end of a simulation experiment would not be included). Individual matrix elements can contain sets of values based on the nature (or absence) of the interaction (described below). For the purposes of operating on the model parameterization for inference (e.g., model calibration), the MRM represents the set of allowed/forbidden interactions: this both constrains the size of the possible search space and prevents the automated model exploration from positing solutions that lead the model to generate the desired behavior, but are forbidden (i.e., due to natural laws) in the real-world. So, while a matrix element with a value of "0" would indicate the absence of the representation of the interaction represented by the matrix element, a 'null' (or other marker) value would indicate that the interaction is forbidden. We also note that every row in the MRM can be expressed as a propositional statement. A specific example is shown below in Section 2.1.1.
2. Model Rule Structure (MRS): The MRS contains the nature of each individual interaction present in the potential model. The MRS annotates each row of the MRM with the functional form in the interaction between relevant variables. Thus, the synthesis of the MRM and MRS can comprehensively represent functional relationships present in the system (equations for a differential equation model or rules for an agent-based model) and be operated upon using MLME techniques for prognostic or diagnostic inference. To increase the representational capacity of a given simulation model, the SSKR will be allowed to contain alternate (but real-world plausible) functional forms, as it is possible that the exact manner in which the model variables interact is unknown (e.g., the numerous forms of energy potential used in nuclear structure calculations), but several are plausible based on real-world data. Should this be the case it may be desirable to consider these alternate functional interactions when calibrating the model or exploring parameter space. We note again that it is not required that alternate functional forms be specified in the MRS; it is an option to allow for enhanced representational capacity of the model/simulation. A specific example is shown in Section 2.1.1. The MRS uses the Mathematical Markup Language (MathML)[6, 7] as this is a generalizable markup language which can describe both the structure and content of mathematical expressions or equations in machine-readable and human-readable formats.
3. The discretization, dimensionality, and topology (DDT) of the model world. The dimensionality, size, and nature of the space (discrete/level of discretization, continuous, network) must be specified. This is of paramount importance for the automated (or assisted) conversion of a model to an executable simulation model, however it can also be critical for physics-based models that only seek to approximate reality (e.g., one- or two-dimensional approximations of a quantum system or a supernova). We anticipate that a user would select from a pre-specified list of options concerning spatial dimensionality (1D, 2D, 3D), discretization (discrete, continuous) for space/time as relevant, level of discretization (fixed or adaptive) for space/time as relevant, world boundary conditions (Dirichlet, Neuman, etc.), and spatial structure information (i.e., network, contiguous, etc.). All options, with the exception of spatial structure information, can be selected from a list. The spatial structure information can be read in manually (if applicable) and used by the CMA to generate simulation code (See section 2.2.2). The list of options and a specification for defining topologies can be expanded for spatially-explicit/spatially-complex models.

4. Model Flow Matrix (MFM): The MFM is the operations specification of the model components. To convert the model into an executable simulation model, one must specify the order of events in the model and the fundamental unit of repetition, that is, the highest-level operation through which the model iterates; this could be a time step, sum, spatial calculation, etc. The operational specification is represented with a directed graph (stored as an adjacency matrix), with each node of the graph representing either a function (i.e., row on the MRM with associated data element in the MRS) or a sub-MFM; in this sense, the MFM can be considered a nested object, with the level of nesting dependent on the nature of the system being modeled. This representation is necessary for ABMs and Discrete Event Simulations (DES) in which the order of events, or the order in which the agents perform their actions, can be important.
5. Model Knowledge Matrix (MKM): The MKM holds the real-world knowledge that was used to characterize that interaction. This is an itemized list of human-readable knowledge, which is an output of whatever IE tool is employed. Each element in an MRM matrix element-tuple will be associated with the one or more indices in this itemized list which contain information that was used to characterize or develop that interaction in the model.

The SSKR is stored in a Pandas dataframe [8], as this allows for the storage of multiple variable types in a single data structure and is readily able to handle the (if necessary) nested elements of the SSKR model specification.

2.1.1 Model Representation Example

For illustrative purposes, we consider the Epidemiological model, Bucky [9]. Bucky is a multi-compartment epidemiological model which calculates the number of Susceptible, Exposed, Infected, and Recovered (SEIR) individuals in the face of a pandemic. This model was chosen as an example for demonstrating the SSKR because it is complex and relates to a field, epidemiology, that has been of considerable recent interest due to the COVID-19 pandemic. Also, our research group does not focus on epidemiological modeling, and therefore we believe the characterization of Bucky provides an informative example for demonstrating the generalizability of MAGCC's workflow. First, we will demonstrate how the SSKR and its different components would be utilized for model representation, expansion, composition, and decomposition. We will be presenting the example on a single spatial/age compartment of the Bucky model, as Bucky divides the United States spatially at the county level and divides population according to age. The MRM presented below is for a single county/age (denoted by the subscripts I and j). but is sufficient to illustrate the method. We have extracted the following model mathematical framework from the Bucky code:

1) $\frac{dS_{ij}}{dt} = -\beta_{ij} S_{ij} \left( I_{ij}^a + I_{ij}^m + I_{ij}^h \right)$
2) $\frac{dE_{ij}}{dt} = \beta_{ij} S_{ij} \left( I_{ij}^a + I_{ij}^m + I_{ij}^h \right) - \sigma E_{ij}$
3) $\frac{dI_{ij}^a}{dt} = (1 - \alpha) \sigma E_{ij} - \gamma I_{ij}^a$
4) $\frac{dI_{ij}^m}{dt} = \alpha (1 - \eta_i \upsilon_j) \sigma E_{ij} - \gamma I_{ij}^m$
5) $\frac{dI_{ij}^h}{dt} = \alpha \eta_i \upsilon_j \sigma E_{ij} - \gamma I_{ij}^h$
6) $\frac{dR_{ij}^h}{dt} = \gamma I_{ij}^h - \tau R_{ij}^h$
7) $\frac{dR_{ij}}{dt} = \gamma \left( I_{ij}^m + I_{ij}^a \right) + (1 - \phi_i) \tau_i R_{ij}^h$
8) $\frac{dD_{ij}}{dt} = \phi_i \tau_i R_{ij}^h$

Where $1/\gamma$ represents the infectious period, $\phi_i$ represents the case fatality rate for age group i, $\tau_i$ represents the recovery period from infection for age group i, $\alpha$ represents the proportion of cases that are asymptomatic, $\eta_i$ is the fraction of cases needing hospitalization, $1/\sigma$ represents the viral latent period, and $\beta_{ij}$ represents the probability of infection for age group i at location j; we note that this matrix is a complex object and can be considered to be a sub-model. For simplicity/illustration, we only consider the fundamental model (i.e., in a single age/location compartment). Bold elements in equations 4 and 5 are discussed in section 2.1.2. The MRM is shown in Table 1.

| MRM | S | E | $I^a$ | $I^m$ | $I^h$ | $R^h$ | R | D |
|---|---|---|---|---|---|---|---|---|
| dS/dt | $-\boldsymbol{\beta_{ij}}$ | 0 | 1 | 1 | 1 | 0 | 0 | 0 |
| dE/dt | $\boldsymbol{\beta_{ij}}$ | $\sigma$ | 1 | 1 | 1 | 0 | 0 | 0 |
| $dI^a$/dt | 0 | $\alpha, \sigma$ | $\gamma$ | 0 | 0 | 0 | 0 | 0 |
| $dI^m$/dt | 0 | $\alpha, \sigma, \eta_i$ | 0 | $\gamma$ | 0 | 0 | 0 | 0 |
| $dI^h$/dt | 0 | $\alpha, \sigma, \eta_i, \tau_i$ | 0 | 0 | $\gamma$ | 0 | 0 | 0 |
| $dR^h$/dt | 0 | 0 | 0 | 0 | $\gamma$ | $\tau_i$ | 0 | 0 |
| dR/dt | 0 | 0 | $\gamma$ | $\gamma$ | 0 | $\tau_i, \phi_i$ | 0 | 0 |
| dD/dt | 0 | 0 | 0 | 0 | 0 | $\tau_i, \phi_i$ | 0 | 0 |

Table 1: Bucky MRM

To maintain concordance with real-world/logical constraints, it will be denoted that the numerical constants cannot vary and altering a parameter for a specific row will alter that parameter's value for all other rows where it is present. The MRS would then be:

1) $-c_1^1(c_3^1 + c_4^1 + c_5^1)$
2) $c_1^2(c_3^2 + c_4^2 + c_5^2) - c_2^2$
3) $(1 - c_{2,1}^3)c_{2,2}^3 - c_3^3$
4) $c_{2,1}^4(1 - c_{2,3}^4)c_{2,2}^4 - c_4^4$
5) $c_{2,1}^5 c_{2,3}^5 c_{2,2}^5 c_{2,4}^5 - c_5^5$
6) $c_5^6 - c_6^6$
7) $c_3^7 + c_4^7 + (1 - c_{6,2}^7)c_{6,1}^7$
8) $c_{6,1}^8 c_{6,2}^8$

Where $c_{1,2}^3$ indicates that you would multiply the second defined parameter (singlet in the potential matrix element-tuple) in the third row and first column of the MRM by the variable/quantity represented by that column. There is no explicit spatial configuration for this model, as spatial interactions are approximated by the matrix, $\boldsymbol{\beta_{ij}}$.

The final piece of information necessary to complete the SSKR is the provenance of the knowledge that was used to construct the model. To illustrate how we envision this, consider a matrix of the same dimensionality as the MRM. Each matrix element can then be a list of integers, each of which is an index pointing to an entry in a list. For example, the matrix element in the third row and second column of the MRM is $\alpha$. The sibling entry in the MKM could hypothetically be:

[2,3], pointing to the list:
2. $\alpha$ is the fraction of infections that are asymptomatic.
3. $\alpha$=0.37.

### 2.1.2 Model Comparison Example

The combination of the MRM and MRS are sufficient to compare models. In some circumstances, mathematical operations can perform the same function but be presented completely differently. As an example, consider $\sin(x)$ vs. $\frac{e^{ix}-e^{-ix}}{2i}$. To a naïve modeler, these might appear to be different expressions, while in actuality they are just different ways to express the same mathematical operation. To account for this, MRM/MRS pairs are compared with each other using a system designed to test pairs of functions against each other (e.g., one set of functions was extracted from code, one set was extracted from documentation using NLP) by randomly sampling the functions (with identical variables) over their respective domains. Should the functions return identical output for all random samples we consider them to equivalent; in the event that the functions return different outputs, they will be inspected by a domain expert (human) to identify the reason for the difference (i.e., is this a mistake?) and posit a solution.

### 2.1.3 Model Extension/Composition Example

The overall structure of the Bucky model is shown in Fig. 2. The model as it is currently constructed considers three classes of Infected (motivated by COVID-19): asymptomatic, mild, and hospitalized. A natural extension of this model could be the addition of an Intensive Care Unit compartment, from which a patient could either recover or die, and to which the hospitalized class could progress in addition to the 'Recovered From Hospital' compartment. We illustrate this hypothetical addition in Fig. 3.

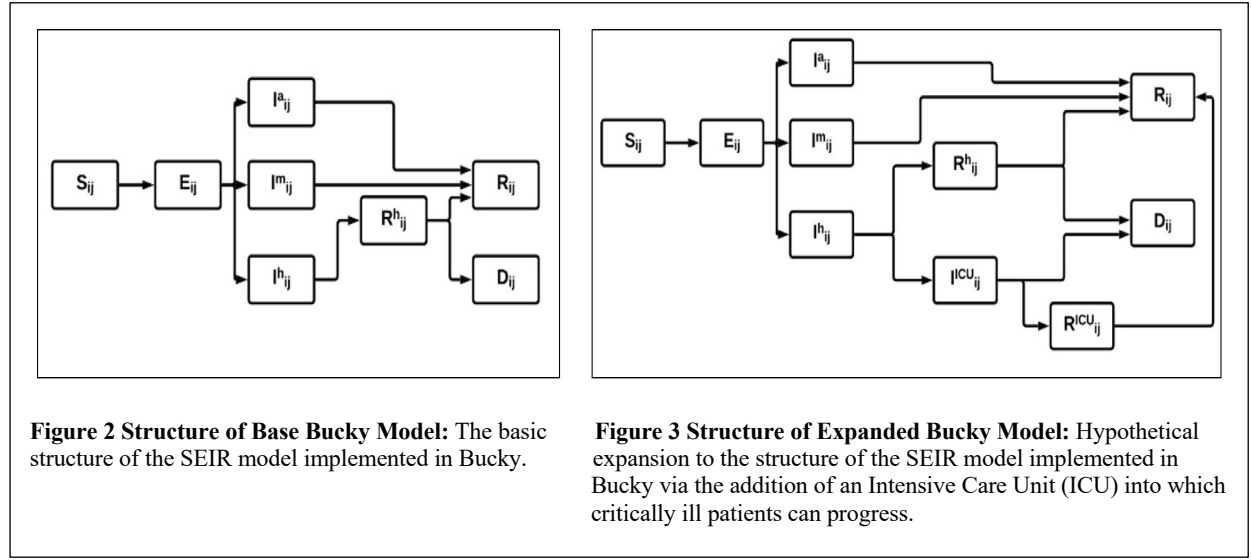

**Figure 2 Structure of Base Bucky Model:** The basic structure of the SEIR model implemented in Bucky.

**Figure 3 Structure of Expanded Bucky Model:** Hypothetical expansion to the structure of the SEIR model implemented in Bucky via the addition of an Intensive Care Unit (ICU) into which critically ill patients can progress.

This requires the addition of two equations and the modification of six:

$$\frac{dS_{ij}}{dt} = -\beta_{ij}S_{ij}\left(I_{ij}^a + I_{ij}^m + I_{ij}^h + I_{ij}^{ICU}\right); \frac{dE_{ij}}{dt} = \beta_{ij}S_{ij}\left(I_{ij}^a + I_{ij}^m + I_{ij}^h + I_{ij}^{ICU}\right) - \sigma E_{ij};$$

$$\frac{dI_{ij}^h}{dt} = \alpha\eta_i\sigma E_{ij} - \gamma I_{ij}^h - \xi_i I_{ij}^h; \frac{dR_{ij}^h}{dt} = \gamma I_{ij}^h - \tau_i R_{ij}^h;$$

$$\frac{dR_{ij}}{dt} = \gamma\left(I_{ij}^m + I_{ij}^a\right) + (1-\phi_i\tau_i)R_{ij}^h + (1-\varphi_i\psi_i)R_{ij}^{ICU}; \frac{dD_{ij}}{dt} = \phi_i\tau_i R_{ij}^h + \varphi_i\psi_i R_{ij}^{ICU};$$

$$\frac{dI_{ij}^{ICU}}{dt} = \xi_i I_{ij}^h - \gamma I_{ij}^{ICU} - \phi_i I_{ij}^{ICU}; \frac{dR_{ij}^{ICU}}{dt} = \gamma I_{ij}^{ICU} + (1-\psi_i)R_{ij}^{ICU}$$

Where we have introduced the infected and recovered ICU compartments and introduced the parameters $\varphi_i$ to represent the ICU case-fatality rate for age group i, $\xi_i$ to represent the fraction of hospitalized cases that necessitate admission to the ICU, and $\psi_i$ to represent the recovery period from an infection that required admission to the ICU for age group i. The MRM for the extended model is shown in Table 2.

| MRM | S | E | $I^a$ | $I^m$ | $I^h$ | $I^{ICU}$ | $R^h$ | $R^{ICU}$ | R | D |
|---|---|---|---|---|---|---|---|---|---|---|
| dS/dt | $-\beta_{ij}$ | 0 | 1 | 1 | 1 | 1 | 0 | 0 | 0 | 0 |
| dE/dt | $\beta_{ij}$ | σ | 1 | 1 | 1 | 1 | 0 | 0 | 0 | 0 |
| d$I^a$/dt | 0 | α, σ | γ | 0 | 0 | 0 | 0 | 0 | 0 | 0 |
| d$I^m$/dt | 0 | α, σ, $\eta^i$ | 0 | γ | 0 | 0 | 0 | 0 | 0 | 0 |
| d$I^h$/dt | 0 | α, σ, $\eta^i$, $\tau^i$ | 0 | 0 | γ, $\xi^i$ | 0 | 0 | 0 | 0 | 0 |
| d$I^{ICU}$/dt | 0 | 0 | 0 | 0 | $\psi^i,\xi^i$ | γ, $\varphi^i$ | 0 | 0 | 0 | 0 |
| d$R^h$/dt | 0 | 0 | 0 | 0 | γ, $\xi^i$ | 0 | $\tau^i$ | 0 | 0 | 0 |
| d$R^{ICU}$/dt | 0 | 0 | 0 | 0 | 0 | γ | 0 | $\psi^i$ | 0 | 0 |
| dR/dt | 0 | 0 | γ | γ | 0 | 0 | $\tau^i,\phi^i$ | $\psi^i,\varphi^i$ | 0 | 0 |
| dD/dt | 0 | 0 | 0 | 0 | 0 | 0 | $\tau^i,\phi^i$ | $\psi^i,\varphi^i$ | 0 | 0 |

Table 2: Extended Bucky MRM

The MRS would then be:

1) $-c_1^1(c_3^1 + c_4^1 + c_5^1 + c_6^1)$
2) $c_1^2(c_3^2 + c_4^2 + c_5^2 + c_6^1) - c_2^2$
3) $(1 - c_{2,1}^3)c_{2,2}^3 - c_3^3$
4) $c_{2,1}^4(1 - c_{2,3}^4)c_{2,2}^4 - c_4^4$
5) $c_{2,1}^5 c_{2,3}^5 c_{2,2}^5 c_{2,4}^5 - c_{5,1}^5 - c_{5,2}^5$
6) $c_{5,1}^6 c_{5,2}^6 - c_{6,1}^6 - c_{6,2}^6$
7) $c_{5,1}^7 - c_7^7 - c_{5,2}^7$
8) $c_6^8 - c_8^8$
9) $c_3^9 + c_4^9 + (1 - c_{7,2}^9)c_{7,1}^9(1 - c_{8,2}^9)c_{8,1}^9$
10) $c_{7,1}^{10} c_{7,2}^{10} + c_{8,1}^{10} c_{8,2}^{10}$

The flow matrix is only necessary for the implementation of the expanded model into code, so is not part of a model-extension task. Lastly, any new knowledge that was used to develop the model extension would be added to the MKM.

2.1.4 Model Decomposition Example

We interpret 'model decomposition' to refer to the process of identifying features/elements of the model that can be separated and cast as a (more-complex) sub-model; this would allow for higher-fidelity representation of selected components of the model, depending on the use-case, and is therefore integral to the sustainment and customization of models. In the Bucky model, the probability that a susceptible person becomes exposed is based on a single matrix element taken from a matrix that is designed to represent this probability as a function of age and location as well as other factors (e.g., degree of non-pharmaceutical intervention at that location for that age range). As such, this single parameter is actually representative of a more complex model consisting of a spatial interaction network and various factors that can influence viral transmission dynamics. In our hypothetical example, we decompose the primary model of viral transmission dynamics (the SEIR model specified above), and the spatial interaction network by positing a more complex model to represent/calculate this parameter. For illustration, consider this sub-model to be an Agent-Based Model (ABM) consisting of a set of human agents, each of whom have the SEIR compartment operating internally.

In that case, the MRM columns would be extended to include $\beta_{ij}$ and any variables which determine the value of $\beta_{ij}$. The matrix elements for these columns would be 'null,' as in this example, those variables would only interact with the viral transmission compartment through the $\beta_{ij}$ parameter, which is calculated externally by a sub-model. Additionally, a new set of rows, defining the rule structure of the ABM governing the spatial location of individual humans, would be concatenated onto the bottom of the old MRM. Thus, the decomposition/expansion of the model will be clearly evident by the structure of the MRM. The MRS would be extended to incorporate rules that govern the movement of the human agents.

In this model decomposition, elements of the SKR that were previously un-necessary (MFM and DDT) will become critically important. First, there is both an explicit representation of space (through which the agents move) and there will necessarily be some topology (e.g., agents move from unique homes, through shared public transportation, to shared work or grocery store, and back to unique home) which constricts the movement of the agents. We would likely define this as a set of connected two-dimensional grids, with each location having a unique grid, and connected with a network structure. Further, as the model is now no longer a set of inter-connected differential equations, the flow-diagram of the model becomes important, and this specifies when to calculate the $\beta_{ij}$ parameter (as a function of distance and location) and update the agent's internal state (i.e., S, E, I [and sub-types], R, or D). A visualization of this MFM can be seen in Figure 4, where entities (domain specific and identified by the CMA) that are looped over are indicated in bold and the functions that correspond to the MRM/MRS rows are italicized (see Table 1).

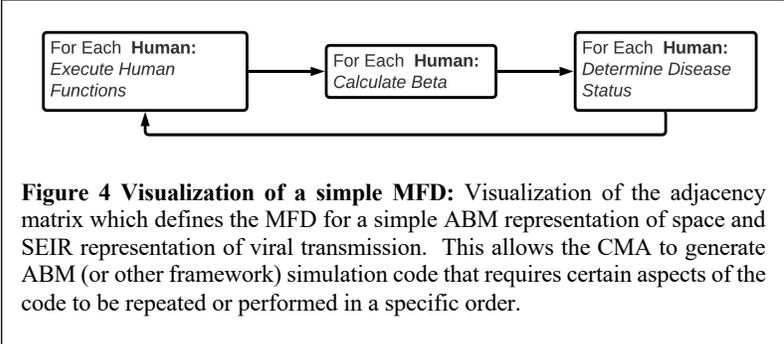

**Figure 4 Visualization of a simple MFD:** Visualization of the adjacency matrix which defines the MFD for a simple ABM representation of space and SEIR representation of viral transmission. This allows the CMA to generate ABM (or other framework) simulation code that requires certain aspects of the code to be repeated or performed in a specific order.

*2.2 Machine-Automated Model Composition using the Computational Modeling Assistant (CMA)*

There are numerous ways to represent knowledge mathematically as a model, however not all knowledge is representable with all types of models. The CMA [10-13] takes as input propositional/logic statements (or, more fundamentally, rows from the MRM), and returns what types of models can be constructed with the provided knowledge and data, or what knowledge/data is needed to construct a model of a specified type (e.g., ABM). An overview of the CMA's capabilities is shown in Figure 5. The construction of models by the CMA is specified as a planning task. Artificial intelligence planning is broadly defined as: given an initial state and a desired goal, discover the sequence of actions or rules that will transform that initial state to the final state which satisfies the goal. That sequence of actions is called the plan, and there may be many plans that fulfill the objective or no plans. The basic elements of a planning task include 1) representation of states, 2) representation of goals, 3) representation of actions, 4) representation of plans, and 5) an algorithm to discover one or more plans.

The CMA utilizes and leverages developments in formal knowledge representation and AI by using an intelligent-agent approach based upon a logical framework. It performs automated reasoning to aid in the construction of a computational model (and ultimately simulation code), treating model construction as a planning task where the goal is translating a collection of knowledge structures into simulation code. The CM''s knowledgebase of modeling and simulation is independent of any scientific domain and can be extended with new modeling techniques and mathematical methods as needed. The CM''s logical framework is based on rewriting logic, Maude [14], that is simple yet expressive enough to support multiple scientific domains. It provides capabilities such as reflection and formal verification beyond other logical frameworks such as CLIPS or Prolog. These additional capabilities are useful for the CMA as it can provide alternative models and analyze the knowledge database for inconsistencies or gaps. The logic is Turing-complete with no limits on representational power and provides capabilities such as reflection (higher-order and temporal logics), model checking, formal verification, and inference strategies. The CMA constructs and represents models with a flexible logical formalism that can be instantiated with a variety of computational frameworks for execution. As the CMA develops plans for the specification of models from the SSKR these different modeling techniques correspond to alternative plans: thus there is not necessarily a single representation of a model composed from the SSKR but rather numerous potential representations based on the researchers requirements and/or the comprehensiveness of the completion of the fields of the SSKR. The CMA provides these alternatives to the researcher such that one or more can be chosen for implementation. The logical basis of the CMA gives it numerous capabilities within an automated scientific extraction and modeling framework such as composition of models from simpler models, traceability of the model construction process, model comparison at the domain or structural level, and model checking for completeness and identify gaps. The CMA uses a standard breadth-first search as an inference algorithm that selects mapping rewrite rules to be applied to the set of Maude logical statements, and incrementally constructs the model specification as each mapping rewrite rule is applied. The algorithm creates a state transition graph where goal states correspond to complete model specifications, and a path from the initial state to a particular goal state is the sequence of mapping rewrite rules for transforming the conceptual biological model into a model specification. While Maude uses breadth-first search as the standard algorithm, its meta-level reflection capabilities allow strategies to be defined that control the rewriting inference process. Such strategies are written as part of a mapping rewrite rule, and the CMA can use strategies for high-order reasoning that cannot be performed with first-order logical inference. Furthermore, because the exact sequence of rules can be recovered, this provides an explanatory description to the user about how the biological model was transformed into a computational model, which might be useful for pedagogical or debugging purposes.

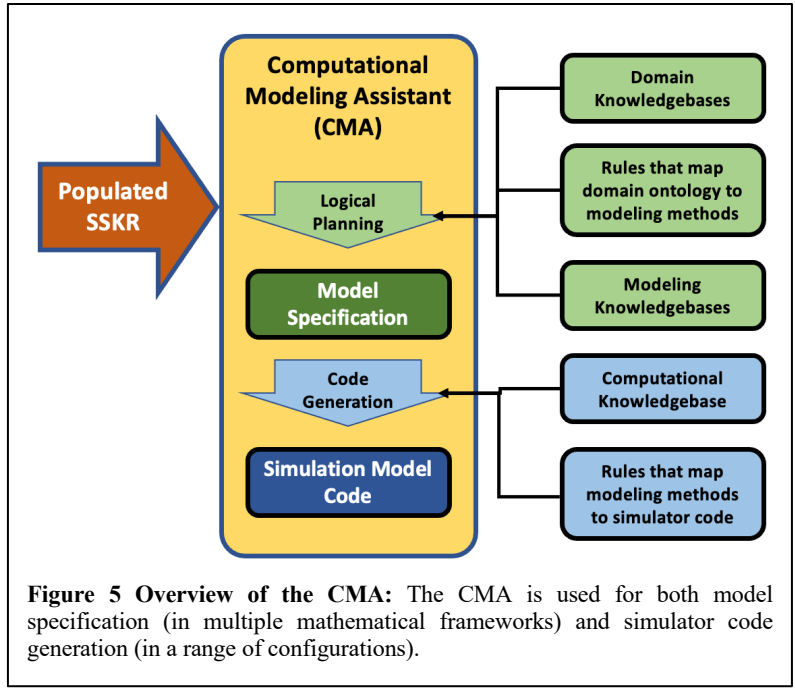

**Figure 5 Overview of the CMA:** The CMA is used for both model specification (in multiple mathematical frameworks) and simulator code generation (in a range of configurations).

This ability to provide an explanatory description of the sequence of inference rules is important in the application of the CMA to real-world circumstances. We have noted above that the SSKR is intended to have comprehensive expressiveness in terms of all the potential forms of models it can represent, however not every modeling task will necessarily require every field of the SSKR to be populated. When the CMA interfaces with the SSKR it receives logic statements based on the specific content of a specific instance of the SSKR; this represents the initial state of the planning task. The goal state of the planning task is a model specification where all of the statements from the SSKR have been accounted for in the list of potential computational models. If the desired modeling method is reached, then the planning task is successful and the CMA proceeds with code generation for a specific simulation model based on which of the available methods is desired for the particular use. However, planning fails if:

1. There are one or more logic statements from the SSKR that could not be transformed into a modeling method. This could be due to the lack of a mapping rule for that particular domain specific term into a specific mathematical or modeling/simulation form. In this case the CMA returns the point at where the planning fails and informs the user that additional information is needed to completely represent the knowledge model that was transformed into the SSKR.
2. There are model specifications that cannot be reached from the initial state defined by a non-comprehensively populated SSKR. In this case the CMA returns the set off modeling methods NOT reachable based on the content of the SSKR, and, through the planning trace process noted below, can identify what additional information would need to be inputted into the SSKR if such a modeling method were desired for use.

### 2.2.1 CMA Biomedical Example

To demonstrate how the CMA, at its current level of development, can use structured knowledge to generate a model specification, and then use that model specification to generate code/an executable simulation model (see section 3.1.1), we will consider a model of gut mucus stratification, modeled at the cellular/molecular level [10].

This is an example demonstrating the current capabilities of the CMA where we consider an auto-regulatory gene network consisting of a gene, G, that can be transcribed into mRNA, R, which can decay and be translated into Protein, P, and form a dimer, P2, that positively promotes the transcription of G through promoter, PR. This conceptual model can be instantiated with a set of natural language statements:

1. *P2* positively regulates *G*
2. *G* is transcribed into *R*
3. *R* is translated into *P*
4. *R* degrades
5. *P* degrades
6. *P* dimerizes to form *P2*
7. *P2* dissociates into *P*

The CMA performs a series of steps that includes 1) syntactically and semantically parsing each statement, 2) determination of the words (typically nouns) in each statement that are the variables in the system, 3) the determination of the words (typically verbs) in each statement that are the processes, 4) the construction of a model specification for the derivative of each variable, 5) the transformation of each process into a model specification for a mathematical function, and 6) casting the derivate for each variable to the set of mathematical functions that effects that variable to form a model specification for a set of ODEs.

Currently, the CMA categorizes the variables and processes in the biological statements during the task of parsing those statements by annotating each word to ontologies of biological concepts. The CMA utilizes existing biomedical ontologies such as the Gene Ontology [15, 16] Systems Biology Ontology (SBO) [17], and the Foundational Model of Anatomy[18] (See Table 3).

| Entity | Reference | Process | Reference |
|---|---|---|---|
| *G* is a **gene** | SBO:0000243 | *positively regulate* | SBO:0000459 |
| *P2* is a **Protein Complex** | SBO:0000297 | *transcribe* | SBO:0000183 |
| *R* is a **messenger RNA** | SBO:0000278 | *translate* | SBO:0000184 |
| *P* is a **polypeptide chain** | SBO:0000252 | *degrade* | SBO:0000179 |
| *R,P* is a **material entity** | SBO:0000240 | *dimerize* | SBO:0000177 |
| Table 3: Using knowledgebases to define domain-specific model terms | | *dissociate* | SBO:0000180 |

The set of natural language statements then becomes:

1. **protein complex** *positively regulates* **gene**
2. **gene** *is transcribed into* **messenger RNA**
3. **messenger RNA** *is translated into* **polypeptide chain**
4,5. **material entity** *degrades*
6. **polypeptide chain** *dimerizes* to form **protein complex**
7. **protein complex** *dissociates into* **polypeptide chain**.

This ontological annotation provides a semantic categorization of the entities and processes in the conceptual biological model that will need to be transformed into a computational model. Since our objective in this example is to focus on the transformation process of biological concepts into computational constructs, we purposefully constrain our near-natural language parsing capabilities to avoid issues such as ambiguous meanings, entity name recognition, and other issues associated with free-form natural language. Using a constrained language will not limit the type or scope of biology that can be modeled; rather, it will require researchers to describe their hypotheses in an unambiguous standard format. Our expectation is that in the future we can interface the CMA with more advanced information extraction systems that would extract modeling statements directly from literature, or to NLP tools that would allow free-form language expression.

With the set of annotated statements in this example, the CMA performs logical reasoning to construct a model specification. The CMA executes actions (apply mapping rules) to transform each biological statement into a portion of the model specification, and the task is complete when all of the biological statements have been transformed. The sequence of actions performed by the agents is the resultant plan, the steps to transform a biological model (initial state) into a computational model (goal state). The CMA is currently able construct a Ordinary Differential Equation (ODE) model, a Partial Differential Equation (PDE) model, a Petri net, or certain primitive ABMs (See Table 4).

| Rule | Model Implementation |
|---|---|
| X **messenger RNA** *is translated into* Y **polypeptide chain** | ODETerm(Y,H(x)) |
| X **material entity** *degrades* | ODETerm (X,-kX) |
| X **polypeptide chain** *dimerizes to* Y **protein complex** | ODETerm(X,-kX) **AND** ODETerm(Y,kX) |
| X **protein complex** *dissociates into* Y **polypeptide chain** | ODETerm(X,-kX) **AND** ODETerm(Y,kX) |

Table 4: parsed natural language to equations

In the presented example the CMA successfully produced a hybrid ODE/PDE model of gut mucus stratification [10] with the biochemical network represented by an ODE inside each cell and the extracellular mucus stratification layers represented by a PDE. The parsed statements and resulting equations are shown in Table 5.

| CMA-Parsed Statement | Equation |
|---|---|
| translate(muc2-mRNA, muc2) | $\frac{d[muc2_{mRNA}]}{dt} = H(muc2_{gene})$ |
| transcribe(muc2-gene, muc2-mRNA) | $\frac{d[muc2]}{dt} = H(muc2_{mRNA}) - S(muc2)$ |
| transcribe(A-gene, A-mRNA) | $\frac{d[A_{mRNA}]}{dt} = H(A_{gene})$ |
| translate(A-mRNA, A-protein) | $\frac{d[A]}{dt} = H(A_{mRNA}) - B(A,B)$ |
| transcribe(B-gene, B-mRNA) | $\frac{d[B_{mRNA}]}{dt} = H(B_{gene})$ |
| translate(B-mRNA, B-protein) | $\frac{d[B]}{dt} = H(B_{mRNA}) - B(A,B)$ |
| bind(A-protein, B-protein, AB-complex | $\frac{d[AB]}{dt} = B(A,B) - S(AB)$ |
| secrete(goblet, muc2, colon, muc2[e]) | $\frac{d[muc2_E]}{dt} = S(muc2)$ |
| diffuse(AB-complex[e]) dissociate(AB-complex[e], A-protein[e], B-protein[e] | $\frac{d[AB_E]}{dt} = \nabla^2 AB_E + S(AB) - D(AB_E)$ |
| diffuse(A-protein[e]) decay(A-protein[e]) | $\frac{d[A_E]}{dt} = \nabla^2 A_E + D(AB) - k_1 A_E$ |
| diffuse(B-protein[e]) decay(B-protein[e]) | $\frac{d[B_E]}{dt} = \nabla^2 B_E + D(AB) - k_2 B_E$ |

**Table 5: CMA generated hybrid ODE/PDE model of gut mucus stratification.**

2.2.2   CMA Automated Generation of Simulation Model Code from Model Specifications

The current version of the CMA is able to generate executable code for the models specified through its planning capability. For the example of gut mucus dynamics the CMA also developed the system of equations (a DE model) by linking specific mathematical equations (e.g., Hill function, exponential decay) to their associated variables (see Figure 6) [10].

```
reactions = (
        // lasI/lasR/rsaL quorum sensing
        //
        {type=signal; name=r20; products=(lasI); reactants=(LASR_N3OXHSL_dimer,RSAL); rate=Hill;}, // regulation
        {type=signal; name=r21; products=(lasI); reactants=(lasI); rate=MassAction;}, // mRNA decay
        {type=signal; name=r22; products=(lasI); reactants=(); rate=MassAction;}, // basal transcription
        {type=signal; name=r25; products=(LASI); reactants=(lasI); rate=Hill;}, // translation
        {type=signal; name=r26; products=(LASI); reactants=(LASI); rate=MassAction;}, // protein degradation
```

**Figure 6 System of equations code specification generated by CMA:** Defines the types of equations that are informed by or update variables in the model.

Generation of simulation code requires specifying the solver to be used (as this is a differential equation model), the compute platform (GPU in this case), and any instrumentation required for simulation exploration (e.g., calibration to data, parameter sweeps, perturbations, Bayesian inference). The CMA annotates the model specification with this additional information and requests the code generation module, specifically BioSwarm, to produce source code files (see Figure 7).

```
__device__ void
hpi_1_kernel_function(int ODEnum, int speciesNum, int pitch, float *speciesData, float *speciesData_in,
float *speciesData_out, float *speciesParameters, float mh, float mf)
{
    float interactionValue, termValue, F_val;
    int rIndex, rNum, rType;
    int rsIndex, rSpecies;
    int pIndex;

    // check knockout
    if (hpi_1_struct->speciesFlags[speciesNum*hpi_1_struct->idx_intPitch+ODEnum]) {
        speciesData_out[speciesNum*pitch+ODEnum] = 0;
        return;
    }

    // loop through each reaction for species
    interactionValue = 0.0;
    rIndex = 0;
    rNum = hpi_1_struct->speciesReactions[speciesNum*hpi_1_struct->idx_srPitch+rIndex];
    while (rNum >= 0) {
        rType = hpi_1_struct->reactionType[rNum];
        pIndex = 0;
        switch (rType) {

            case 0:
                // Hill function
                rsIndex = 0;
                rSpecies = hpi_1_struct->reactionTable[rNum*hpi_1_struct->idx_rPitch+rsIndex];

                termValue = 1.0;
                while (rSpecies >= 0) {
                    F_val = speciesData[rSpecies*pitch+ODEnum] + mf * speciesData_in[rSpecies*pitch+ODEnum];
                    termValue *= SHILL(F_val, PARAM_PMAX, PARAM_C, PARAM_H);

                    // next reactant
                    ++rsIndex;
                    pIndex += 3;
                    if (rsIndex > hpi_1_struct->maxReactants) rSpecies = -1;
                    else rSpecies = hpi_1_struct->reactionTable[rNum*hpi_1_struct->idx_rPitch+rsIndex];
                }

                interactionValue += termValue;
                break;
```

**Figure 7: Executable simulator code generated by CMA**

The source code needs to be compiled with the appropriate compilers, linkers, and libraries on the destination compute platform to produce the final executable file to be run. These annotations are parameters that are external to the model and utilized by EMEWS for simulation management and exploration. For example, a simulation model specific implementation of the $2^{nd}$-order Runge-Kutta finite-difference method (see Figure 8) has a delta time step (dt) parameter.

```
//
// Gene network models
//
geneModel = {
    type = DynamicReactionRuleModel;
    name = PseudomonasQS;
    debug = NO;

    // finite difference method
    numericalScheme = RungeKutta2ndOrder;
    //numericalScheme = AdaptiveRungeKuttaFehlberg;
    ratio = 1;
    dt = 0.05;

    maxReactions = 32;
    maxReactants = 32;

    species = (amet, anth, Pi,
               lasR, LASR, lasI, LASI, N3OXHSL, LASR_N3OXHSL, LASR_N3OXHSL_dimer, rsaL, RSAL,
               rhlR, RHLR, rhlI, RHLI, C4HSL, RHLR_C4HSL, RHLR_C4HSL_dimer, RHLR_N3OXHSL,
               phoB_p, mvfR, MVFR, pqsABCD, PQSA, HHQ, pqsH, PQSH, PQS, MVFR_HHQ, MVFR_PQS);
```

**Figure 8 Simulator Specification code generated by CMA:** Defines the system as a mass-action chemical reaction network, to be solved with an $2^{nd}$ order Runge Kutta solver.

### 2.3 Machine Learning-augmented Model Exploration (MLME) for calibration and model comparison

Once a simulation model has been created it is necessary to calibrate that model to real-world data, validate that model by examination of its behavior, and evaluate the suitability of that model for a specified task (which may involve comparing that model to another simulation model). These tasks can be broadly labled as model exploration (ME), which invariably involves running in silico experiments runs to iteratively examine the simulation model's behavior and capabilities using a series of well-defined processes: 1) executing the simulation model given a set of input parameters reflecting conditions of interest, 2) assessing the simulation model's output and behavior based on metrics of interest, 3) alter the input parameters or underlying assumptions based on some improvement strategy and 4) repeat from Step 1 until some desired criteria are satisfied. As simulation models grow in complexity, it rapidly becomes computationally intractable to apply brute-force methods to large-scale ME. The advent of ML and AI has provided sets of tools that can aid in running ME tasks at scale. Some of these methods include: various active learning (AL)[19, 20] approaches, including those using Gaussian Process (GP) surrogate models [21, 22], and ML models such as Random Forest [23, 24]; approximate Bayesian computation approaches [25, 26], including sequential Monte Carlo [27] and incremental mixture ABC [28]; evolutionary approaches, including single [29] and multi-objective [30] genetic algorithms [5, 31]; and data assimilation approaches, including ensemble Kalman filtering [32] and particle filters[33, 34]. Ongoing development in these methods has also provided a robust ecosystem of various libraries that can provide state-of-the-art ME capabilities, such as Keras [35, 36], scikit-learn [37], scikit-opt [38], and DEAP [39] in Python; mlrMBO [40], randomForest [41], EasyABC [42], IMABC [28], hetGP [43], laGP [44] and caret [45] in R; and TensorFlow [46] bindings that are available in both. The MLME environment is flexible enough to employ this range of tools.

As an example of the work flow for parameter space characterization, MLME employs an integration of two techniques, Genetic Algorithms [47-49] and Active Learning (AL), for the purposes of identifying real-world data plausible regions of parameter space, a crucial step in the calibration of complex simulation models. AL is a sub-field of ML which focuses on determining the optimal selection of training data to be used to train an ANN or statistical model [50], and can be used for classification [51, 52] or regression [53, 54]. AL is especially useful for modeling problems with a large amount of unlabeled data where manually labeling that data is expensive, or if simulation data is being used, precludes the need to run all the simulations needed to fully characterize the behavior space (which may be computationally intractable); a term applied to this use is "metamodeling", for which AL has been used in industrial contexts that utilize complex simulation models [55]. In these circumstances (specifically the costly data labeling) AL provides near comprehensive characterization of model parameter space for a minimal computational cost (in terms of cpu-time), in which AL rendered this problem computationally tractable by reducing the number of simulations require to characterize the search space by ~99% when compared to a brute-force exploration [56]. Note that in this process the ANN is NOT a surrogate of the computational model, but rather a classifier intended to identify patterns of correlation (e.g. the metamodel) between a MRM parameterization and the real-world plausibility (e.g. the ability to encompass the variability seen in the real world data) of that parameterization.

In the MLME workflow for parameter space characterization a GA produces an ensemble of real-world data defined plausible model structure/parameter pairs, each a single point in a high-dimensional space (e.g. the MRM space). AL efficiently determines the real-world data plausible parameter ranges around those points. The output of the AL workflow is an ANN *classifier*, which allows the user to establish whether a specific model structure/parameter pair without executing the model. The classifier takes a parameterized MRM as input and returns one of two possible model parameterization classes: 1) real-world plausible (meaning the computational model would generate data compatible with the experimental/real-world data) or 2) not real-world plausible. This process will be performed separately for each unique MRM structure. To begin, we posit the existence of some function,

$$y = f(\vec{x}), x \in \chi \subset R^n, y \in R,$$

Which accepts as input a parameterized MRM and predicts the associated class, and that this function can be approximated given input data from the training set:

$$D_{train} = \{x_j^t, f(x_j^t)\} \text{ for } j = 1, \dots, n,$$

where $x_j^t$ represents a single MRM used to train the ANN. The ANN model uses a binary cross-entropy loss function [57]. The procedure begins by training, using gradient-descent, an ANN to approximate the function described above, and thus predict model-parameterization class. The initial training dataset uses all of the real-world plausible parameterizations identified by the GA; this dataset is then augmented by random MRM's such that the initial training set will contain balanced class populations. This dataset is then used to train the ML model. The algorithm then ranks the remaining unlabeled parameterizations by class-membership uncertainty (defined in [58]) such that the parameterizations whose class is most uncertain in the current ML model are then selected for labeling and the process repeats until a stopping criterion (sufficient classification accuracy) is reached. This trained ANN then enables the efficient exploration of the high-dimensional model parameter space: each neural net computation can be performed on the order of milliseconds, while it takes on the order of 10 minutes or more to execute a single instance/replicate of the model. The mathematical details of this process can be seen in Ref [56].

The current version of the MLME environment can perform parameter space characterization [56], calibration [5], surrogate model development [59] and control-discovery [31, 60, 61] tasks and is flexible enough to integrate additional ME capabilities and algorithms as needs arise.

## 3.0 Discussion and Summary

MAGCC will dramatically improve the ability to automate model composition and decomposition for their creation, sustainment, and customization and greatly enhance the machine-assisted/automated construction of scientific simulation models to perform prognostic or diagnostic inference. To date, these tasks in the scientific simulation community are almost always performed with bespoke, customized models/simulations, heavily reliant on the continuity of individual programmers/developers, resulting in significant challenges in terms of model hand-off to other users for iterative refinement/updating, concatenation with complimentary models, modification for use, and extraction of the exact knowledge purported to be contained in these models. Furthermore, specific scientific domains or modeling techniques/formalisms present distinct challenges in terms of calibration and parameterization of highly complex simulation models. Given the complexity of these models, such tasks almost invariably involve the application of advanced machine learning (ML) methods for simulation model exploration, of which there are many different techniques and implementations. Therefore, there is an additional layer of interoperability and integration required to provide machine-assistance for these tasks. The innovation of MAGCC is in recognizing that a solution to the challenge of machine-assisted simulation development and use for scientific simulation requires addressing the seemingly paradoxical requirements of generalizable expressiveness and efficient customization; we do this by designing MAGCC around three core components, the SSKR for formal knowledge representation, the CMA for model specification and simulation model composition/implementation, and a suite of ML-enhanced ME capabilities. Each of these components is intrinsically comprehensive in their expressiveness and inclusiveness but is designed for flexibility and expansibility for customized tasks/domains. We arrived at this approach by leveraging our experience developing simulation models in multiple domains (cellular/molecular pathophysiology, low energy nuclear physics, epidemiology, and social science) without being daunted by how complex those simulations may need to be to robustly answer the problems of interest. This modeling approach of embracing complexity (versus being limited by preconceived notions of tractability) has led us to develop robust and scalable means to deal with that complexity. We apply the same philosophy to the design of MAGCC.

Future work on the base MAGCC system will expanding the representational capabilities of all of its components. The interfaces for the SSKR will be augmented by the development of additional scripts to aid in the conversion of the content of its subcomponents into MathML. For the CMA we will be developing a general-purpose knowledge base, using generic terminology (e.g., variable1 causes variable2 to increase) that is motivated by our work developing the domain-specific knowledgebases. Because this is general-purpose, the CMA will be unable to automatically posit the mathematical form of the relationship based on domain knowledge. As such, we will extend the CMA to accept functional forms from the MRS using MathML. To compare the knowledge encoded by a model, e.g., complex mathematical functional formats, we will implement a sampling/comparison scheme into the CMA to establish whether two functions expressed in different ways return the same output. Also, to increase the representation capabilities of th CMA, it will be extended to generate time-delay differential equations, as these are critical to a range of modeling processes. We will also be explicitly considering multi-scale models, implementing the ability of the CMA to specify a set of 'sub-models' that interface in a complex manner. This structure help to inform heterogenous code (e.g., CPU/GPU) development for very-large scale simulation models. Lastly, in addition to our own expanding range of modeling projects (i.e. extending from biomedicine into zoonotic systems, geopolitical modeling of undergoverned spaces and infinite geopolitical contests), we hope that external interest in MAGCC will lead to the development of additional domain-specific knowledgebases that will enhance the utility of the system. We will also extend the code-generating capability of the CMA beyond its current ability to generate differential equation (both ordinary and partial) and discrete event (Petri-Net, Boolean and Network) simulation models with a specific goal of providing a suite of capabilities for agent-based modeling. We will also increase flexibility for users to choose solvers for their differential equation simulations. For systems of ODEs, we will allow for the user-selection and comparison of a suite of solvers, including a suite of Runge-Kutta methods, Adams-Moulton methods, Nystrom methods, and multiderivative methods. For PDEs, we will incorporate the ability to solve basic PDEs with validated solvers (i.e., the heat equation or wave equation); however, we recognize that many PDEs do not have generalizable techniques to efficiently solve them. We will initially focus on solvers designed for MHD equations, including a variety of Riemann methods and Godunov-type methods. Once we are able to handle the PDEs that model space weather, we will then focus on solving arbitrary PDEs. To achieve this, we will implement a generic finite difference scheme for PDE-solving and allow the user to select from a pre-defined list of iteration schemes or manually define their own, depending on the use case. For example, the iteration scheme for the heat equation, $\frac{\partial f(x)}{\partial t} = \frac{\partial^2 f(x)}{dx^w}$, would be $k\frac{f(x-h)-f(x+h)-2f(x)}{h^2}$; a user will manually define any iteration scheme and associated step size (h) using MathML. Finally, we are aware of the interest in the use of AIs for automated code generation [2]. The architecture of the CMA facilitates the collection and structuring of knowledge required for such systems to work, and therefore future work will include the development and incorporation of additional knowledgebases to facilitate the use of newly-developed AI code generators.

With additional development and the above expansion of this capabilities MAGCC will provide a revolutionary leap in the process of developing, updating, assessing for use, and refining simulation models. This capability will be available to every scientific domain that uses simulation through the future-proofed extensibility of the MAGCC. MAGCC has the potential to revolutionize how policy/decision makers interact with information provided by simulation models by providing unprecedented transparency, flexibility and update-capability, all critical features necessary to establishing trust and confidence in such simulations.

**Acknowledgements:** This work was supported in part by the National Institutes of Health Award UO1EB025825. This research is also sponsored in part by the Defense Advanced Research Projects Agency (DARPA) through Cooperative Agreement D20AC00002 awarded by the U.S. Department of the Interior (DOI), Interior Business Center. The content of the information does not necessarily reflect the position or the policy of the Government, and no official endorsement should be inferred.